\documentclass[twocolumn]{autartGIO}    

\usepackage{graphicx}      
\usepackage[numbers]{natbib}        

\usepackage{latexsym}
\usepackage{amsmath}
\usepackage{amsfonts}
\usepackage{epstopdf}

\usepackage{color}

\usepackage[colorlinks=true,linkcolor=blue,urlcolor=blue,citecolor=blue]{hyperref}

\newcommand{\be}{\begin{equation}}
\newcommand{\ee}{\end{equation}}

\newcommand{\bea}{\begin{eqnarray}}
\newcommand{\eea}{\end{eqnarray}}
\newcommand{\bean}{\begin{eqnarray*}}
\newcommand{\eean}{\end{eqnarray*}}

\newcommand{\ac}{\`}

\def\bfa{{\bf a}}

\def\bfr{{\bf r}}
\def\bfs{{\bf s}}

\def\bfv{{\bf v}}
\def\bfu{{\bf u}}
\def\bfx{{\bf x}}
\def\bfy{{\bf y}}
\def\bfz{{\bf z}}

\def\bfI{{\mbox{\boldmath $I$}}}

\def\bfmu{{\boldsymbol \mu}}
\def\bfnu{{\boldsymbol \nu}}

\def\bfomega{{\boldsymbol \omega}}

\begin{document}

\onecolumn
\clearpage
\thispagestyle{empty}

\begin{center}
{\Large 
PRIN project MARIS: Marine Autonomous Robotics for InterventionS,  call of year 2010-2011, prot. 2010FBLHRJ.	
} \\
\rule{0.0cm}{1cm}\\

\noindent 
Research Unit UNISALENTO:  MARIS Technical Report version of  August, 31st 2014: updated version (with minor typing corrections) of the first version of July 29th, 2013.  \\

\noindent 
\rule{0.0cm}{1cm} 
Further results on the observability analysis and observer design for single range localization in {3D} \\

\noindent 
\rule{0.0cm}{1cm} 
Giovanni Indiveri and Gianfranco Parlangeli, \\
Dipartimento Ingegneria dell'Innovazione \\
University of Salento (ISME node), Lecce, Italy\\

\noindent 
\rule{0.0cm}{1cm} 
\end{center}
  
\twocolumn


\begin{frontmatter}

\title{Further results on the observability analysis and observer design for single range localization in {3D}} 

\thanks[footnoteinfo]{This paper was not presented at any IFAC 
meeting. Corresponding author G.~Indiveri. Tel. +39-0832-297220. 
Fax 39-0832-297733.}

\author[Paestum]{Giovanni~Indiveri}\ead{giovanni.indiveri@unisalento.it},    
\author[Paestum]{Gianfranco~Parlangeli}\ead{gianfranco.parlangeli@unisalento.it}                

\address[Paestum]{Dipartimento Ingegneria Innovazione - ISME node \\
   Universit\ac a del Salento, via Monteroni, 73100 Lecce - Italy}  


\date{ }

\begin{abstract}                          
Single range localization in 3D is an open challenge with important implications for service robotics applications. The issue of single range observability analysis and observer design for the kinematics model of a 3D agent subject to a constant unknown drift velocity is addressed. 
The proposed approach departs from alternative ones and leads to the definition of a linear time invariant state equation with a linear time varying output that can be used to globally solve the original nonlinear state estimation problem with a standard linear estimator. Simple necessary and sufficient observability conditions are derived. Numerical simulation examples are reported to illustrate the performance of the method.  
\end{abstract}

\end{frontmatter}
\pagenumbering{arabic} 

\section{Introduction}\label{sec:Problem}
The problem of single range based localization is relevant in several land  \cite{Martinelli:05}, \cite{zhou2008robot}, aerial \cite{kassas2012observability} and marine robotics \cite{Song:99} \cite{Arrichiello:11} \cite{JOE-URA:13} applications. In essence, the problem consists in estimating an agent's position exploiting knowledge about its motion model (typically its kinematics model where the velocity is a known input), a range measurement from a source and eventually other sensor readings related to the vehicle's attitude. The challenge of using single range information for localization is related to the fact that traditional trilateration algorithms used in systems as the Global Positioning System (GPS), long base line (LBL) or ultra short base line (USBL) underwater navigation systems are ill posed when only range from a single source is known. Yet fusing information from a motion model of the agent (including velocity and attitude) and a single range measurement can be sufficient to estimate the position of the agent. Finding the conditions on the agent's motion state that allow to estimate its position from a single range measurement is an observability problem that needs to be tackled in order to eventually design an observer. Given that range is a nonlinear function of the position, even if the motion model of the vehicle should be linear, the observability issue is inherently nonlinear. A major contribution on observability for nonlinear systems is \cite{Hermann} where the fundamental ideas and results about local and weakly local observability are described: single range localization studies building on differential geometric tools need to tackle the difficulties related to local and weakly local observability as opposed to the global observability concept known for linear systems. Such issues are clearly addressed, by example, in references \cite{jouffroy2006algebraic}, \cite{jouffroyremarks} and \cite{Gadre:04}. 

Single range aided localization is particularly relevant in cooperative navigation applications where a team of vehicles needs to perform collaborative motion control tasks while possibly performing relative localization exploiting intra-vehicle communication to measure relative ranges. Such problems arise in several underwater robotics scenarios \cite{Fallon:10} \cite{WeWaWhEu-TRO2013} \cite{SoAgPaMa-ICRA2013} as well as in more general settings \cite{CaYuAn:11}. Moreover, the problem of range based localization is technically similar to the problem of source localization where a vehicle knowing its own position is asked to estimate the position of a target from which it acquires range measurements \cite{DaFiDaAn:09} 
\cite{InPeCuNGCUV2012}. Indeed the localization solution described in this paper was partially inspired by the work in \cite{InPeCuNGCUV2012} and it can be used to address the problem outlined in \cite{DaFiDaAn:09} as illustrated in section \ref{sec:simulations}.     

A milestone contribution in the area of single range localization is given by the work of Batista et al. \cite{Batista:10} \cite{Batista:11}: they propose to study the single range localization problem of an agent subject to a constant, but unknown, drift velocity through an augmented state approach. The original nonlinear system (where the state belongs to $\mathbb{R}^6$ and is made of the agent's position and unknown drift velocity components) is transformed in a linear time varying (LTV) system in $\mathbb{R}^9$ through an augmented state technique. This leads to the remarkable result of allowing to study the {\em global} observability properties of the system with well known Gramian based tools of LTV systems theory \cite{Rugh:96} and of designing a Kalman filter for global state estimation. The LTV system derived in \cite{Batista:10} \cite{Batista:11} is of the form
\bea
&&{\dot \bfz} = A(t,\bfu(t),y(t))\,\bfz + B \,\bfu(t) \label{batistaeq1} \\
&&y(t) = C \,\bfz \label{batistaeq2}
\eea
namely, the system matrix $A(t)$ explicitly depends, among the rest, on the output $y(t)$ that is the range measurement. More precisely, the $A(t)$ matrix is a function of terms proportional to $1/y(t)$: this poses both fundamental as well as implementation issues. From a theoretical perspective, assuming that the output $y(t)$ should be affected by additive noise, the dependency of $A(t)$ from $y(t)$ implies that some of its  entries are stochastic and that the model uncertainty on the state equation (\ref{batistaeq1}) could not be possibly assumed to be only additive as is usually done within the theory of Kalman filtering. As a consequence assuming  additive gaussian noise on the state and output equations (\ref{batistaeq1}) - (\ref{batistaeq2}), the associated Kalman state estimator  is not guaranteed to be optimal in the usual sense. Indeed the numerical examples provided in \cite{Batista:11} confirm that the Kalman filter estimates converge  to the true state variables, but there is no {\em a priori} guarantee that the estimate is optimal in terms of estimate covariance. A second potential difficulty arising from the structure of equations (\ref{batistaeq1}) - (\ref{batistaeq2}) is related to the eventual stability analysis of 
the Kalman filter on equations (\ref{batistaeq1}) - (\ref{batistaeq2}) coupled with a motion controller. Indeed the presence of the output $y(t)$ in the matrix $A(t)$ does not allow to exploit in a straightforward fashion the standard separation principle used within linear systems theory to study the convergence and stability of state estimation filters coupled with feedback controllers.

As for the implementation of the Kalman filter described in \cite{Batista:11}, given the dependency of some entries of $A(t)$ from $1/y(t)$, it is necessary to assume that $y(t) \neq 0$ at all times. Indeed in \cite{Batista:11} it is assumed that $y(t)$ has strictly positive, finite, lower and upper bounds: yet in real application scenarios with unpredictable sensor noise and outliers one would need to pre-filter the output in order to guarantee the absence of numerical issue related to exceedingly small or null $y$ readings.     

In the light of the above observations and inspired by the work in \cite{InPeCuNGCUV2012} and \cite{Batista:11}, this paper describes an alternative approach to address the single range localization problem. As a result, the same problem addressed in \cite{Batista:11} is globally solved by introducing an LTV system of the form
\bea
&&{\dot \bfz} = A \,\bfz + B\,\bfu(t) \label{noieq1} \\
&&y(t) = C(\bfu(t))\,\bfz \label{noieq2}
\eea
namely where the state equation is completely linear time invariant (LTI) and has dimension $8$ rather than $9$. The output equation is still LTV, but has a very simple structure. The proposed method does not build on state augmentation techniques, but rather exploits the structure of the original state equations expressed in an inertial frame as opposed to the body frame formulation used in \cite{Batista:11}. Within the proposed solution, given the LTI nature of the state equation (\ref{noieq1}), the difficulties related to the dependency of $A(t)$ in equation (\ref{batistaeq1}) from $1/y(t)$ are completely removed. Also notice that the output matrix $C(t)$ in equation (\ref{noieq2}) within the solution presented in this paper depends on the input $\bfu(t)$, but not on the output 
$y(t)$: as a consequence an additive measurement noise would not affect the entries of $C(t)$ nor of $A$ and $B$ hence preserving the optimality of a Kalman filter as a state estimator as long as noise is gaussian (and the input $\bfu(t)$ is perfectly known, i.e. noise free). Indeed, as in \cite{Batista:11}, the localization problem can be globally solved with a standard Kalman filter with dimension $8$ instead of $9$. Moreover, given the extremely simple structure of equations (\ref{noieq1}) - (\ref{noieq2}), the observability analysis is extremely simple and it allows to derive necessary and sufficient observability conditions on the agent input (i.e. its velocity). 

The main ideas and methods used to solve the problem are described in section \ref{sec:methods} on a simplified version of the problem, namely in the absence of drift velocity terms as eventually present in underwater applications due to constant and unknown currents. The resulting strategy for single range localization in the absence of currents is addressed in section \ref{sec2}. The complete case of single range localization in the presence of constant and unknown currents is solved in
 section \ref{sec:currents}. Simulation examples are presented and discussed in section \ref{sec:simulations} while conclusions are summarized in section \ref{sec:conclusions}.

\section{Main ideas and methods}\label{sec:methods}
Consider a 3D kinematics vehicle model by the following equations:
\bea
{\dot \bfx} \,&=&\, \bfu \label{xdoteq} \\
y(t) \,&=&\, \bfx^\top \bfx \,=\, \| \bfx(t) \|^2 \label{yeq} 
\eea
being $\bfx(t) \in \mathbb{R}^{3 \times 1}$ the unknown vehicle position (state), $\bfu(t)  \in \mathbb{R}^{3 \times 1}$ its velocity (known input) and $y(t) \in [0,\infty)$ the measured output at time $t$. 

The issue of estimating the state $\bfx(t)$ is ultimately related to identifying the initial state 
\be
\bfx_0 : = \left. \bfx(t)\right|_{t=0}. \label{bfx0eq}
\ee
In particular, with reference to the model in equations (\ref{xdoteq}) - (\ref{yeq}), $\bfx_0$ is related to the systems input and output by the following
\bea
\bfx(t) &=& \bfx_0 + \int_0^{t} \bfu(\tau)d\tau \label{xtsol} \\
y(t) &=& \bfx^\top \bfx  \;=\;
\| \bfx_0 \|^2 + 2 \left(\int_0^{t} \bfu(\tau)d\tau \right)^\top \bfx_0 + \nonumber \\
& &
+ \left\| \int_0^{t} \bfu(\tau)d\tau \right\|^2 \!\!. \label{eqy1}
\eea
Denoting with
\be
\bfI_\bfu(t)  := \int_0^{t} \bfu(\tau)d\tau  = (I_x(t), I_y(t), I_z(t))^\top \in \mathbb{R}^{3 \times 1}
\label{defItempocontinuo}
\ee
equation (\ref{eqy1}) results in 
\bea
&& {\bar y}(t)  :=   \frac{1}{2}\left( y(t) - y(0) - \| \bfI_\bfu(t) \|^2 \right) \label{defybar} \\
&&  {\bar y}(t) = \bfI_\bfu(t)^\top  \bfx_0 
\label{regression}
\eea
where the left hand side of equation (\ref{regression}) is known as well as the time varying vector $\bfI_\bfu(t)$ on the right hand side. Notice that knowledge of $\bfx_0$ corresponds to observability (by definition) as well as to reconstructability as $\bfx(t)$ can be directly calculated through equation (\ref{xtsol}).

Equation (\ref{regression}) allows to directly identify the necessary and sufficient conditions on $\bfu$ for observability (and reconstructability) on a finite time interval. The analysis is performed both in the discrete and continuous time cases.
\subsection{Discrete time case}
The discrete time version of equation (\ref{regression}) leads to a Least Squares (LS) problem as follows: assume that time is sampled at a rate $T_s$ such that
\be
t_k = k\,T_s \;:\; k =1,2,\ldots, n
\label{timesampling}
\ee
and 
\bea
&& {\bar \bfy} := ({\bar y}(t_1), {\bar y}(t_2), \ldots, {\bar y}(t_n))^\top 
\in \mathbb{R}^{n \times 1} \label{defybarravettore} \\
&& H := 
\left(
\begin{array}{ccc}
I_x(t_1) & I_y(t_1) & I_z(t_1) \\
I_x(t_2) & I_y(t_2) & I_z(t_2) \\
\vdots & \vdots & \vdots \\
I_x(t_n) & I_y(t_n) & I_z(t_n) \\
\end{array}
\right) \in \mathbb{R}^{n \times 3} \label{defH}
\eea
then equation (\ref{regression}) evaluated at each sampling time leads to the LS problem
\be
H \bfx_0 = {\bar \bfy}  \label{LSproblem}
\ee
that admits a unique solution  ${\hat \bfx}_0$ iff rank$(H) = 3$, i.e.  
\be
\mbox{rank}(H) = 3 \;\; \Longleftrightarrow \;\;
{\hat \bfx}_0 = (H^\top H)^{-1}H^\top {\bar \bfy}.  
\label{LSsolutionforx0}
\ee
The discrete time case sufficient and necessary observability condition rank$(H) = 3$ requires that $n \geq 3$ (necessary, but not sufficient) and, most important, that the velocity input $\bfu(t)$ generates columns of $H$ in equation (\ref{defH}) that are linearly independent. Stated formally, the above lead to the following result: \\
 
 \noindent
{\bf Statement 1}  - {\em Observability conditions for the discrete time case.}  \\
\noindent Given a discretized version of model (\ref{xdoteq}) - (\ref{yeq}) by sampling time as in (\ref{timesampling}) with $n\geq 3$, the state $\bfx_0$ is observable (and $\bfx(t)$ is reconstructable) if and only if the velocity signal $\bfu$ guarantees that $H \in \mathbb{R}^{n \times 3}$ in equation (\ref{defH}) has rank $3$. \\
%
\noindent 
{\bf Proof of Statement 1}\\
\noindent It follows by standard LS theory. \\

Contrary to linear systems, the observability depends on the input. 
 A similar observability condition can be derived for the problem of localizing the position of a fixed target on behalf of a moving vehicle from single range measurements. Details are discussed in  \cite{InPeCuNGCUV2012}.
 
 \subsection{Continuous time case}
The line of thought for the continuous time case is the same of the discrete time case. With reference to  equation (\ref{regression})  consider  multiplying  both sides by $\bfI_\bfu(t)$ and integrating over a finite interval, namely:
\be
 \int_0^t \bfI_\bfu(\tau) \bfI_\bfu(\tau)^\top \,  \bfx_0 \, d\tau 
 = \int_0^t \bfI_\bfu(\tau)  {\bar y}(\tau) d\tau.
 \label{gra1}
\ee
The left hand side of equation (\ref{gra1}) can be written in terms of the 
the $3 \times 3$ real Gramian matrix
\be
G(t) := \int_0^t \bfI_\bfu(\tau) \bfI_\bfu(\tau)^\top d\tau \;\; \in \mathbb{R}^{3 \times 3}
\label{gramiano}
\ee
and the right hand side in terms of a vector $\bfmu \in \mathbb{R}^{3 \times 1}$
\be
\bfmu(t) := \int_0^t \bfI_\bfu(\tau) {\bar y}(\tau) d\tau 
\label{defmu}
\ee
leading to writing equation (\ref{gra1}) as
\be
G(t) \bfx_0 = \bfmu(t).
\label{regressionecontinua}
\ee
{\bf Statement 2} - {\em Observability conditions for the continuous time case.} \\
\noindent Given model (\ref{xdoteq}) - (\ref{yeq}) the state $\bfx_0$ is observable (and $\bfx(t)$ is reconstructable) if and only if the velocity signal $\bfu$ guarantees that the Gramian $G(t) \in \mathbb{R}^{3 \times 3}$ in equation (\ref{gramiano}) has rank $3$. \\
\noindent
{\bf Proof of Statement 2}\\
\noindent {\em Sufficiency} \\
If $G(t) \in \mathbb{R}^{3 \times 3}$ in equation (\ref{gramiano}) has rank $3$ then 
equation (\ref{regressionecontinua}) implies 
\be
\bfx_0 = \left(G(t) \right)^{-1}  \bfmu(t).
\label{x0soluzionetempocontinua}
\ee
\noindent {\em Necessity} \\
It is proven by absurd. Suppose that $\bfx_0$ is observable and rank$(G(t)) < 3$ in the time interval $[0,t]$. The hypothesis that rank$(G(t)) < 3$ on $[0,t]$ implies that the kernel of $G(t)$ is nonempty. Consider
\be
\bfnu \in \mathbb{R}^{3 \times 1} : \bfnu \neq {\bf 0}, {\dot \bfnu} = {\bf 0} \mbox{ and } G(t) \bfnu = {\bf 0} 
\ee 
in $[0,t]$. Then by definition of $G(t)$ and of the constant $\bfnu$ it follows 
\bea
0 &=& \bfnu^\top G(t) \bfnu =
\int_0^t  \bfnu^\top \bfI_\bfu(\tau) \bfI_\bfu(\tau)^\top \bfnu d\tau = \nonumber \\
&=&\int_0^t \| \bfI_\bfu(\tau)^\top \bfnu  \|^2 d\tau \;\; \Longrightarrow
\mbox{ by norm properties} \nonumber \\
&& \bfI_\bfu(\tau)^\top \bfnu = 0 \;\; \forall \;\; \tau \in [0,t]. 
\eea
The above result implies that $\bfx_0 + \bfnu$ and $\bfx_0$ are two distinct (because $\bfnu \neq {\bf 0}$) solutions of equation (\ref{regression}) (and hence of equation (\ref{eqy1})) on the finite time interval $[0,t]$, i.e. $\bfx_0 + \bfnu$ and $\bfx_0$ are undistinguishable on the finite time interval $[0,t]$ as they produce the same output $y(t)$. This violates the hypothesis that $\bfx_0$ was observable, hence it is proven that if $\bfx_0$ is observable in $[0,t]$ the Gramian $G(t)$ must be full rank in $[0,t]$.    

The above results allow to identify all the motions (i.e. velocity inputs) generating globally observable states. In particular, in the discrete time case the persistently exiting input velocity $\bfu$ assuring that $H$ in equation (\ref{defH}) has full rank will guarantee position observability. Similarly, in the continuous time case, the persistently exiting input velocity $\bfu$ assuring position observability is given by the values making the Gramian (\ref{gramiano}) invertible. These results can be exploited to design optimal movements (i.e. velocity profiles) with reference to criteria such as, for example, the condition number of  the regression matrix $H$ eventually subject to constraints as the infinity norm of $\bfu$. Other measures for optimal estimation inputs are of course also possible in the line of the system identification literature on optimal experimental design.  

Most important and most remarkably, the proposed framework allows to derive a linear system with a time invariant (LTI) state equation and a time varying (LTV) output equivalent to the one in equations (\ref{xdoteq}) - (\ref{yeq}) on which to design a Kalman filter for solving the localization problem in case that the proposed observability conditions are met. Contrary to the state augmentation approaches discussed in \cite{Batista:11} and exploited also in \cite{PaPeInMCMC2012}, the proposed approach does not require state augmentation and leads to a linear system with a LTI state equation and a LTV output. 

Notice also that the Gramian full rank condition in statement 2 closely resembles the integral  condition of theorem 2.1 in  \cite{DaFiDaAn:09}; indeed, $\bfI_\bfu(t)$ can be viewed as a filtered version of $\bfx(t)$ in the assumption that $\bfx_{t=0} = {\bf 0}$.

\section{Single range localization} \label{sec2}
With reference to equation (\ref{xtsol}), consider the following
\bea
\bfx_0  &=& \bfx(t)  -  \int_0^{t} \bfu(\tau)d\tau =  \bfx(t) - \bfI_\bfu(t)
\label{modello1modificato} \\
\bfx_0^\top \bfx_0 &=&y(0) =  \nonumber \\
&=& y(t) + \| \bfI_\bfu(t) \|^2 - 2\bfI_\bfu(t)^\top\bfx(t)
\eea
that can be re-written as follows
\be
{\bar y}(t) = \bfI_\bfu(t)^\top\bfx(t) 
\label{new-output}
\ee
where the output signal  ${\bar y}(t)$ is the known quantity
\be
{\bar y}(t) = \frac{1}{2}\left[ y(t) - y(0) + \| \bfI_\bfu(t) \|^2 \right].
\label{new-output-eq}
\ee
Having introduced the time varying output equation (\ref{new-output-eq}) the original system in equations (\ref{xdoteq}) - (\ref{yeq}) can be equivalently represented by the LTV  system
\bea
&& {\dot \bfx} = \bfu \label{xdoteq-2} \\
&& {\bar y}(t) = \bfI_\bfu(t)^\top\bfx(t) \label{yeq-2}
\eea 
that is (globally) observable as long as the conditions in Statement 2 are met. The localization problem of estimating $\bfx(t)$ in the model (\ref{xdoteq-2}) - (\ref{yeq-2}) can thus be solved resorting to a Kalman-Bucy filter. Notice that contrary to the standard LTV setting, the output matrix $C(t) = \bfI_\bfu(t)^\top$ in equation (\ref{yeq-2}) 
is a function of the very input, hence observability (as already discussed) will depend on the velocity input $\bfu(t)$. 

The derived continuous LTV system (\ref{xdoteq-2}) - (\ref{yeq-2}) can be discretized through sampling at times $t_k = k T_s$ as in equation (\ref{timesampling}). Assuming that $\bfu(t)$ is constant in any time interval $[kT_s,(k+1)T_s]$, the discrete time version of the LTV system (\ref{xdoteq-2}) - (\ref{yeq-2}) results in
\bea
&&\bfx_{k+1} = \bfx_k + \bfv_k \label{xdoteq-2d} \\
&&{\bar y}_k = \bfI_k^\top \; \bfx_k \label{yeq-2d}
\eea
where $ \bfv_k := \bfu(kT_s)\,T_s,\;  \bfI_k := \bfI_\bfu(kT_s),\;  \bfx_k := \bfx(kT_s),\;  
 {\bar y}_k := {\bar y}(kT_s),\;  y_k := y(kT_s) \;$ and 
\bea
&&{\bar y}_k = \frac{1}{2}\left[ y_k - y(0) + \| \bfI_k \|^2 \right]
\label{defybark}\\
&&\bfI_{k+1} = \bfI_k + \bfv_{k+1} \;\;:\;\; \bfI_0 = {\bf 0}.  \label{Irecursion}
\eea
Given the model structure in equations (\ref{xdoteq-2d}) - (\ref{yeq-2d}), a linear state estimation filter can be designed to estimate the position $\bfx_k$ as long as the observability conditions in Statement 1 are satisfied. An example of such application is outlined and tested in section \ref{sec:simulations}.

\subsection{Persistently exiting velocity input: an example}
Following the design presented in \cite{InPeCuNGCUV2012}, consider the following input velocity $\bfu = (u_1, u_2, u_3)^\top$
\be
u_i(t) = A_i\,n_i\,\omega\,\cos(n_i \omega t)\label{desired-speed}
\ee
where $n_i$ is a positive integer such that $n_i \neq n_j$ for any $i,j \in \{1,2,3\}$, $A_i$ is a design parameter such that $|A_i| \,n_i \,\omega$ is the maximum speed in direction $i$ and $\omega$ is a design parameter chosen as 
\be
\omega = 2\pi / T_0 \;\;:\;\; T_0 = n_0 T_s 
\label{desired-speed2}
\ee 
for some positive integer $n_0$ being $T_s$ the sampling time. With the above designed velocity input, the regression matrix $H$ in equation (\ref{defH}) over a time interval $[0,(n_{t}n_0 T_s)]$ for some positive integer $n_t$ results \cite{InPeCuNGCUV2012} 
to be diagonal greatly simplifying the issue of designing optimal maneuvers for single range localization.   
\section{Single range localization in the presence of constant and unknown currents}
\label{sec:currents}
The described approach can be extended to the case where the kinematic point mass vehicle is subject to a constant and unknown ocean current, namely the same model used in  \cite{Batista:11}. For the sake of comparing the present solution to the one in \cite{Batista:11}, a similar notation will be used. In particular consider an inertial frame $\{I\}$ and a body fixed frame $\{B\}$ such that the SO$(3)$ rotation matrix from frame $\{B\}$ to frame $\{I\}$ is denoted as 
$^{I}\!R_{B}$. Such rotation matrix, as in  \cite{Batista:11}, is assumed to be known. 
The components of a vector $\bfa \in \mathbb{R}^{3}$ in frames $\{I\}$ or $\{B\}$ will be denoted as ${^I}\bfa$ and ${^B}\bfa$ respectively. Following standard kinematics results, the angular velocity of the vehicle satisfies 
\be
^{I}\!{\dot R}_{B} = S(^{I}\!\bfomega_{B/I}) \, ^{I}\!R_{B} =   {^{I}}\!R_{B} \, S(^{B}\!\bfomega_{B/I})  
\ee  
where $S(\bfa) \in \mathbb{R}^{3 \times 3}$ is the skew symmetric matrix associated to the vector product $\bfa\, \times$ and $^{I}\!\bfomega_{B/I}, {^{B}}\!\bfomega_{B/I}$ are the angular velocities of frame $\{B\}$ with respect to frame $\{I\}$ expressed in frames $\{I\}$ and $\{B\}$ respectively. The term ${^{B}}\!\bfomega_{B/I}$ is assumed to be measured by the agent itself thanks to the on board navigation system (Attitude and Heading Reference System  - AHRS). The term $^{I}\!\bfomega_{B/I}$ is also assumed to be accessible as it is given by
\be
^{I}\!\bfomega_{B/I} = {^{I}}\!R_{B} \, {^{B}}\!\bfomega_{B/I}.
\ee
The position of the vehicle is given by $\bfx \in \mathbb{R}^{3 \times 1}$ 
and the fixed landmark in frame $\{I\}$ from which the agent can measure its euclidean distance has fixed position $\bfs  \in \mathbb{R}^{3 \times 1}$ (${^I}{\dot \bfs} = {\bf 0}$). The velocity  $\dot \bfx$ of the agent is given by the superposition of an ocean current (assumed constant in frame $\{I\}$) denoted with $\bfv_f$ and a relative (w.r.t. the water) velocity $\bfv_r$ such that
\bea
^{I}{\dot \bfx} =  {^{I}}\bfv_f + {^{I}}\bfv_r =  {^{I}}\bfv_f + {^{I}}\!R_{B} \,{^{B}}\bfv_r
\label{defxdot}
\eea 
where the relative velocity in the agents frame ${^{B}}\bfv_r$ is assumed to be actively controlled and measured by the agents on board navigation system (typically thanks to a DVL  - Doppler velocity logger - sensor). The agent measures the euclidean distance to the target $\bfs$, namely the norm (or, equivalently, the square norm) of $\bfr$ given by
\bea
{^I}\bfr &:=& {^I}\bfs - {^I}\bfx  \label{defr}\\
{^B}\bfr &=& {^{B}}\!R_{I} \left( {^I}\bfs - {^I}\bfx \right) \\
{^B}{\dot \bfr} &=& 
- {^B}\bfomega_{B/I}(t) \times {^B}\bfr(t) -{^{B}}\bfv_f(t) - {^{B}}\bfv_r(t)  
\label{modelloPorto}
\eea 
where the angular velocity property ${^B}\bfomega_{I/B} = - {^B}\bfomega_{B/I}$ has been used. Equation (\ref{modelloPorto}) together with
\bea
^{B}{\dot \bfv}_f &=& - {^B}\bfomega_{B/I}(t) \times {^B}\bfv_f(t)  \label{modelloPorto2} \\ 
r(t) &=& \| {^B}\bfr(t) \|  \label{modelloPorto3}
\eea
are exactly the same in equation (1) of  \cite{Batista:11} where the localization problem is formulated as a state estimation issue where the state variables are $(^{B}\bfr^\top, ^{B}\bfv_f^\top)^\top$ having dynamics given by equations (\ref{modelloPorto}) - (\ref{modelloPorto2}) and the output map is equation (\ref{modelloPorto3}). In particular, by estimating the term ${^B}\bfr$, the absolute position of the vehicle in the inertial frame can be recovered through equation (\ref{defr}) as
\be
{^I}\bfx = {^I}\bfs - {{^I}\!}R_{B} \, {^B}\bfr.
\ee 
This formulation of the single range localization problem has the advantage of making explicit use of the agent velocities $\bfv_r$ and $\bfomega_{B/I}$ in the body frame where they are actually measured (i.e. $\{B\}$). The drawback is that in body frame $\{B\}$ the current isn't constant. Given that the rotation matrix from the body frame to the inertial frame is nevertheless necessary to recover the absolute position of the agent and that it is thus assumed to be known, one might just as well formulate the localization problem in the inertial frame where the dynamics equations are simpler. Notice, moreover, that in the very approach of \cite{Batista:11} a Lyapunov transformation based on ${^I}\!R_{B}$ is used to transform model   
(\ref{modelloPorto}) - (\ref{modelloPorto3}) in a different dynamical system for observability analysis and observer design. 

In order to exploit the same approach derived in section \ref{sec2} for the current free case,  the system model will be derived in the inertial frame $\{I\}$. In particular from equations (\ref{defxdot}) and (\ref{defr}) it follows that
\bea
{\dot \bfr}\;     &=& -\bfv_f  -  \bfv_r \label{modelloMio} \\
{{\dot \bfv}_f} &=& {\bf 0} \label{modelloMio2} \\
y &=& \| \bfr \|^2 \label{modelloMio3}
\eea
where all vectors are expressed in frame $\{I\}$ and the left hand side superscript $\{I\}$ has been omitted for the sake of notation compactness. Indeed, from this point on, all vectors expressed in the inertial frame $\{I\}$ will be denoted without left hand side superscript $\{I\}$. The term $\bfv_r$ in equation (\ref{modelloMio}) is assumed known as it may be computed as $\bfv_r = {^{I}\!}R_{B}\, {^B}\bfv_r$ given that ${^B}\bfv_r$ is generated by the agents propulsion system and is measured on board as already mentioned.

Consider the integral of equation (\ref{modelloMio})
\bea
\bfr(t)- \bfr_0 &=& -\bfv_f\,t  -  \int_0^t \bfv_r(\tau) d\tau = \nonumber \\
&=&  -\bfv_f\,t  -  \bfI_{\bfv_r}(t) \label{intr} 
\eea
having defined $\bfI_{\bfv_r}(t) \in \mathbb{R}^{3 \times 1}$ as 
\be
\bfI_{\bfv_r}(t)  := \int_0^t \bfv_r(\tau)\, d\tau  
\ee
and
\[
\bfr_0 := \left. \bfr(t) \right|_{t=0}.
\]
Equation (\ref{intr}) allows to compute  
\[
\left(\bfr(t) + \bfI_{\bfv_r}(t) \right)^\top\left(\bfr(t) +\bfI_{\bfv_r}(t)  \right) =  
\left(\bfr_0 - \bfv_f\,t  \right)^\top
\left(\bfr_0 - \bfv_f\,t  \right)
\] 
implying 
\bea
\| \bfr(t) \|^2 + \| \bfI_{\bfv_r}(t) \|^2 +
2\, \bfI_{\bfv_r}^\top(t) \bfr(t)  = \nonumber \\
=\| \bfr_0 \|^2 + \| \bfv_f \|^2 \, t^2 - 
2\, ( \bfr_0^\top\bfv_f)t  
\eea
namely
\bea
\| \bfr(t) \|^2 - \| \bfr_0 \|^2 
+ \| \bfI_{\bfv_r}(t) \|^2 = \nonumber \\
= - 2\, \bfI_{\bfv_r}^\top(t) \bfr(t)   
-2\, ( \bfr_0^\top\bfv_f)t   + 
\| \bfv_f \|^2 \, t^2.
\label{eq71}
\eea
Notice that the left hand side of equation (\ref{eq71}) is made of all known terms and it can be used as a new output map
\bea
{\bar y}(t) &=&
\| \bfr(t) \|^2 - \| \bfr_0 \|^2 
+ \| \bfI_{\bfv_r}(t) \|^2 \nonumber \\
&=&y(t) - y_0 + \| \bfI_{\bfv_r}(t) \|^2
\label{eq:barycurrentcase}
\eea
and the right hand side of equation (\ref{eq71}) can be expressed as a linear time varying (LTV) term in the new state variable 
$\bfz \in \mathbb{R}^{8 \times 1}$
\be
\bfz = (\bfr^\top, ( \bfr_0^\top\bfv_f), \| \bfv_f \|^2, \bfv_f^\top )^\top,
\label{defz}
\ee
i.e.
\bea
{\bar y}(t) &=& C(t) \, \bfz = \nonumber \\
&=&
\left[- 2\, \bfI_{\bfv_r}^\top(t) \;\;\; -2\,t  \;\;\;\;    t^2  \;\;\;\; 0_{1 \times 3}\right]\, \bfz.
\label{newoutput}
\eea
Given the definition of $\bfz$ in equation (\ref{defz}) and the model (\ref{modelloMio}) - (\ref {modelloMio2}), its dynamic equation is linear time invariant (LTI):
\be
{\dot \bfz} = A \, \bfz + B \,\bfv_r
\label{eqdynz}
\ee
namely 
\bea
{\dot \bfz} &=&
\frac{d}{dt}
\left(
\begin{array}{c}
\bfr \\
( \bfr_0^\top\bfv_f) \\
\| \bfv_f \|^2 \\
\bfv_f
\end{array} 
\right) = \nonumber \\
&=&
\left[
\begin{array}{cccc}
0_{3 \times 3} & 0_{3\times 1} & 0_{3\times 1} & -I_{3 \times 3} \\
0_{1 \times 3} & 0 & 0 & 0_{1 \times 3} \\
0_{1 \times 3} & 0 & 0 & 0_{1 \times 3} \\
0_{3 \times 3} & 0_{3\times 1} & 0_{3\times 1} & 0_{3 \times 3} 
\end{array} 
\right] 
\left(
\begin{array}{c}
\bfr \\
( \bfr_0^\top\bfv_f) \\
\| \bfv_f \|^2 \\
\bfv_f
\end{array} 
\right) + 
\nonumber \\
&+&
\left[
\begin{array}{c}
-I_{3 \times 3}  \\
0_{1 \times 3}  \\
0_{1 \times 3}  \\
0_{3 \times 3} 
\end{array} 
\right] \, \bfv_r.
\rule{0.3cm}{0.0cm}
\label{eqLTInew}
\eea
The range-only localization problem of estimating $\bfr$ and the current velocity $\bfv_f$ from a measurement of $\| \bfr \|^2$ in equations (\ref{modelloMio}) - (\ref{modelloMio3}) is hence reduced to a state estimation problem on a linear time invariant state equation (\ref{eqdynz}) - (\ref{eqLTInew}) with an LTV output map (\ref{newoutput}), namely
\be
\left\{
\begin{array}{l}
{\dot \bfz} = A \, \bfz + B \,\bfv_r \\
 {\bar y}(t) = C(t) \, \bfz.  
\end{array}
\right.
\label{eqdynzbis}
\ee
 The LTI state equations, moreover, have a very simple structure. As anticipated 
 in section \ref{sec:Problem}, notice that this results is similar to the one described in \cite{Batista:11}, but with a few significant differences: the state matrix $A$  does not depend on the output and is actually LTI rather than LTV. Only the output map is time varying and depends on the vehicle's velocity.  Moreover the state vector has dimension $8$ rather than $9$. As a consequence, the Gramian observability matrix to be used for observability analysis has a simpler structure as well as the resulting observer that can be chosen to have a Kalman filter structure.  

Estimating $\bfz$ will result in estimating both $\bfr$ and the current velocity $\bfv_f$. From $\bfr$ the absolute position of the vehicle can be computed as  $\bfx = \bfs - \bfr$ (refer to equation (\ref{defr})).

\subsection{Observability analysis in the presence of currents}
The observability properties of system (\ref{eqdynzbis}) can be studied through the observability Gramian
\be
G(t) = \int_0^t e^{{A^\top} \tau}\,C^\top(\tau)\,C(\tau)\,e^{A\tau} d\tau.
\label{gramianoconcorrenti}
\ee
Given the structure of the $A$ matrix in equation (\ref{eqLTInew}), notice that $A^2=0_{8 \times 8}
$ implying that the exponential matrix $\exp(At)$ is simply 
\be
e^{At} = I_{8 \times 8} + At
\label{expA}
\ee
such that $C(t)\exp(At)$ results in
\be
C(t)\,e^{At} = \left[-2\,\bfI_{\bfv_r}^\top(t) \;\;\; -2\,t  \;\;\;\;    t^2  \;\;\;\; 2t\,\bfI_{\bfv_r}^\top(t) \right]
\ee
and  
\bea
&&\exp({A^\top}\!\tau)\,C^\top(\tau)\,C(\tau)\,\exp(A\tau)= \nonumber \\
=
&&\left[
\begin{array}{cccc}
4\,\bfI_{\bfv_r} \,\bfI_{\bfv_r}^\top & 4\tau\,\bfI_{\bfv_r} & -2\tau^2\,\bfI_{\bfv_r} &-4\tau\,\bfI_{\bfv_r}\bfI_{\bfv_r}^\top \\
4\tau\,\bfI_{\bfv_r}^\top & 4\,\tau^2  &   -2\tau^3  & -4\tau^2\,\bfI_{\bfv_r}^\top \\
-2\tau^2\,\bfI_{\bfv_r}^\top  & -2\,\tau^3 &  \tau^4  & 2\tau^3\,\bfI_{\bfv_r}^\top \\
-4\tau \bfI_{\bfv_r}\,\bfI_{\bfv_r}^\top  & -4\tau^2\bfI_{\bfv_r} &  2\tau^3\bfI_{\bfv_r}  & 4\tau^2\bfI_{\bfv_r} \,\bfI_{\bfv_r}^\top 
\end{array}
\right] \nonumber \\
&& \label{argG}
\eea 
where the dependency of $\bfI_{\bfv_r}$ from $t$ has been omitted for the sake of notation compactness. 

As for the observability conditions, following standard results for LTV systems \cite{Rugh:96}, the model in equation (\ref{eqdynzbis}) will be completely observable in the time interval $[0,t]$ if and only if the Gramian given by equations (\ref{gramianoconcorrenti}) and (\ref{argG}) has full rank. Moreover, the structure of equation  (\ref{argG})  implies that a necessary condition for the complete observability of (\ref{eqdynzbis}) in the time interval $[0,t]$ is that 
\be
G_{11}(t):=4\int_0^t \bfI_{\bfv_r}(\tau) \,\bfI_{\bfv_r}^\top(\tau) d\tau 
\label{eqG11}
\ee
has full rank, i.e. three. Notice that this latter necessary condition for the observability in the presence of constant and unknown currents was shown to be both necessary and sufficient in the current free case (Statement 2). Overall, the observability properties in the presence of constant currents can be summarized as follows. \\

\noindent 
{\bf Statement 3} - {\em Observability conditions for the continuous time case with constant current.} \\
\noindent The model in equations (\ref{newoutput}) - (\ref{eqLTInew}) is observable on $[0,t]$ if and only if the velocity signal $\bfv_r$ guarantees that the Gramian in equations (\ref{gramianoconcorrenti}) and (\ref{argG}) has full rank. Moreover a necessary condition for full observability on $[0,t]$ is that the matrix $G_{11}(t) \in \mathbb{R}^{3 \times 3}$ in equation (\ref{eqG11}) has rank $3$. \\
\noindent
{\bf Proof of Statement 3}\\
The necessary and sufficient conditions on the Gramian in equations (\ref{gramianoconcorrenti}) and (\ref{argG}) follow from standard LTV systems theory \cite{Rugh:96}. As for the necessary condition on the rank of the matrix $G_{11}(t) \in \mathbb{R}^{3 \times 3}$ in equation (\ref{eqG11}), following the same method used to prove Statement 2 it results that if $G_{11}(t)$ should {\em not} be full rank on $[0,t]$, there would exist a constant vector $\bfnu \in \mathbb{R}^{3 \times 1},\, \bfnu \neq {\bf 0}$ such that 
$\bfI_{\bfv_r}(\tau)^\top \bfnu = 0 \;\; \forall \;\; \tau \in [0,t]$: this implies that any vector parallel to 
$\bfz^* = (\alpha\,\bfnu^\top,\, 0,\, 0,\, \beta\,\bfnu^\top)^\top \in \mathbb{R}^{8 \times 1}$ for any constant $\alpha, \beta \in \mathbb{R}$ would belong to the kernel of the Gramian (\ref{gramianoconcorrenti}) - (\ref{argG}) that, hence, would not be full rank. This proves that rank$(G_{11}(t)) = 3\; : \; G_{11}(t) \in \mathbb{R}^{3 \times 3}$ is defined in equation (\ref{eqG11}) is a necessary condition for the observability in $[0,t]$ of the model in equations (\ref{newoutput}) - (\ref{eqLTInew}). \\

Notice that if the $\bfv_r$ components should be defined as in equations (\ref{desired-speed}) - (\ref{desired-speed2}), the observability conditions of Statement 3 would indeed be satisfied. 
Also notice that the results in Statement 3 allow to consider optimal design issues of the vehicle's input $\bfv_r$: in Kalman filtering theory, in fact, the observability Gramian is related to the estimate covariance and to the Fisher information matrix. Building on the results in Statement 3, one could formulate optimal design problems for the input $\bfv_r$ aiming at maximizing, by example, metrics as the norm or the determinant, or the condition number of the resulting Fisher information matrix.   

\section{Kalman filter design and simulation examples}
\label{sec:simulations}
With reference to the model in equations  (\ref{xdoteq-2}) - (\ref{yeq-2}) derived from the original system  (\ref{xdoteq}) - (\ref{yeq}), assume to explicitly account for noise in the standard Kalman filtering framework, i.e. consider the system
\bea
&&\bfx_{k+1} = \bfx_k + \bfv_k  + \bfomega_k \label{xdoteq-2dnoise} \\
&&{\bar y}_k = \bfI_k^\top \; \bfx_k  + \varepsilon_k \label{yeq-2dnoise}
\eea
where $\bfomega_k$ and $ \varepsilon_k $ are zero mean state and output mutually independent disturbances with covariances
\bea
\mbox{cov}(\bfomega_k) &=& E\left[\bfomega_k \bfomega_k^\top\right] = Q_k \\ 
\mbox{cov}(\varepsilon_k)  &=& E\left[\varepsilon_k^2 \right] = R_k.
\eea 
Denoting with ${\hat \bfx}_{k|k}$ the Kalman estimate at step $k$ and with ${\hat \bfx}_{k+1|k}$ the model prediction, the localization Kalman filter results in
\bea
&& {\hat \bfx}_{k+1|k} = {\hat \bfx}_{k|k} + \bfv_k  \label{prediction} \\
&& P_{k+1|k} = P_{k|k} + Q_k \label{predizionecov} \\
&&\bfI_{k+1} = \bfI_k + \bfv_{k+1} \;\;:\;\; \bfI_0 = {\bf 0}  \label{Irecursion-bis}\\
&& K =\left(P_{k+1|k}^{-1} + \bfI_{k+1} \, R_{k+1}^{-1} \, \bfI_{k+1}^\top \right)^{-1}\bfI_{k+1} \, R_{k+1}^{-1} \\
&& {\hat \bfx}_{k+1|k+1} = {\hat \bfx}_{k+1|k} + K ({\bar y}_{k+1} - \bfI_{k+1}^\top \, {\hat \bfx}_{k+1|k} )
\eea
where $P_{k+1|k}$ is the covariance of the prediction ${\hat \bfx}_{k+1|k}$ while
\be
\mbox{cov}({\hat \bfx}_{k+1|k+1}) = \left(P_{k+1|k}^{-1} + \bfI_{k+1} \, R_{k+1}^{-1} \, \bfI_{k+1}^\top \right)^{-1}.
\label{covarianza-stima-Kalman}
\ee 

\begin{figure}[thf]	
\begin{center}
\includegraphics[width=8.0cm]{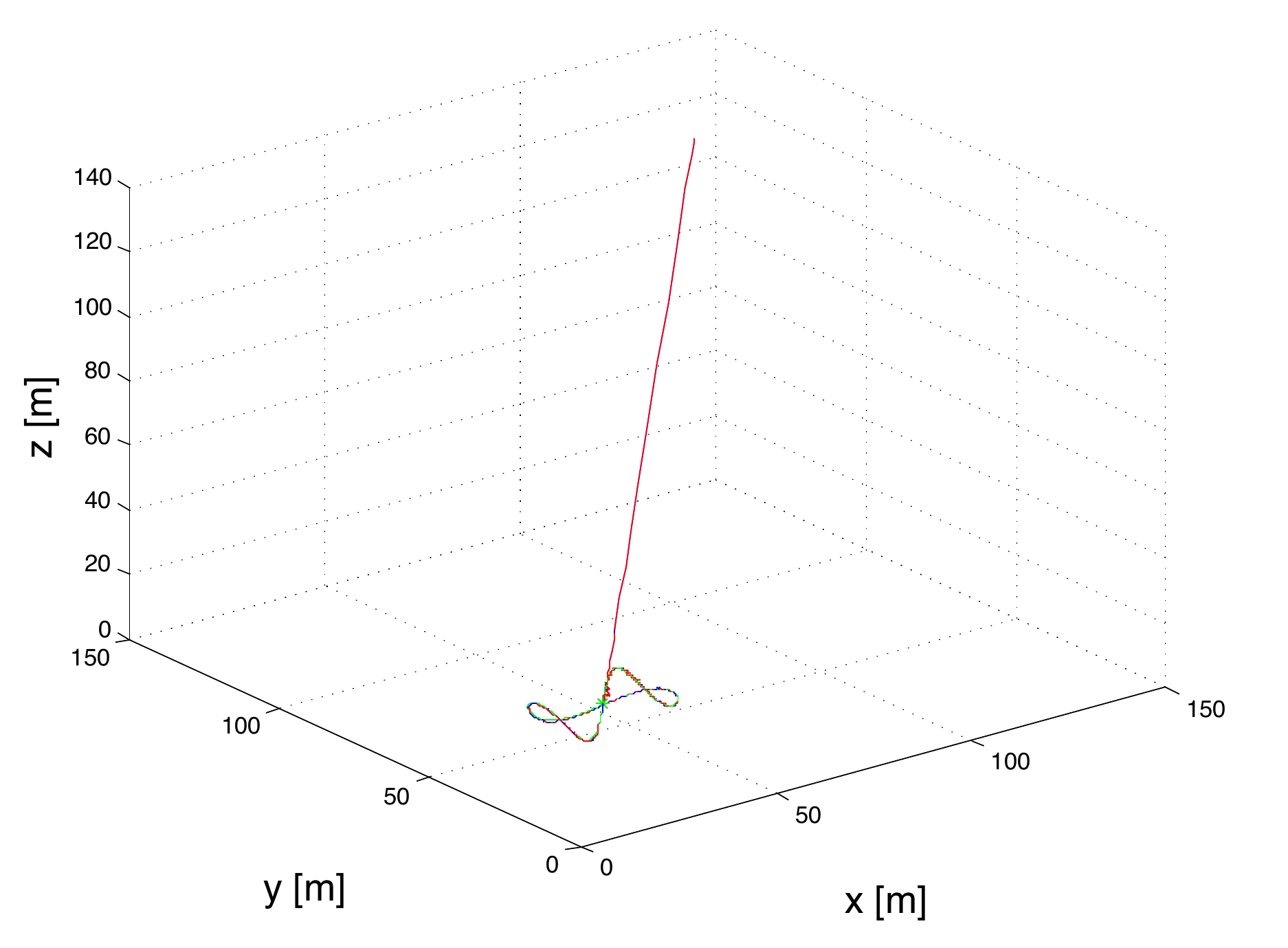}    
\includegraphics[width=8.0cm]{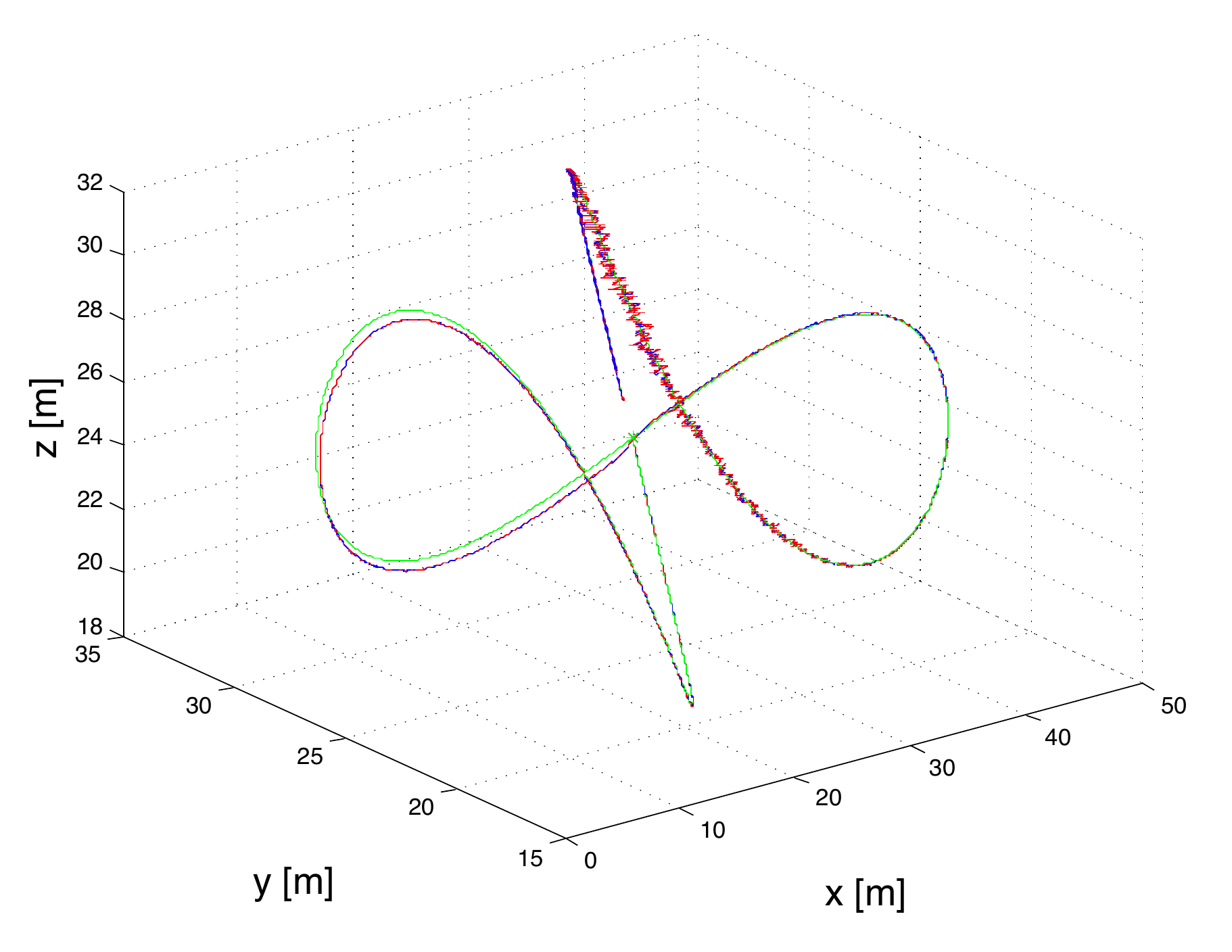}    
\caption{Kalman filter estimated position in the current free case. Refer to the text for details.} 
\label{figtrajnocurrent}
\end{center}
\end{figure}

The proposed Kalman filter described in equations (\ref{prediction}) - (\ref{covarianza-stima-Kalman})  for the current free case  has been implemented and tested using inputs of the form described in equations (\ref{desired-speed}) - (\ref{desired-speed2}). The simulation parameters relative to equations  (\ref{desired-speed}) - (\ref{desired-speed2}) and (\ref{prediction}) - (\ref{covarianza-stima-Kalman}) are summarized in the following table: 
\begin{center}
    \begin{tabular}{| c | c | c |} 
    \hline
    Parameter & Value \rule{0cm}{0.01cm} & Dimension \\ \hline
    $\bfx_0$ & $(25,25,25)^\top$ \rule{0cm}{0.01cm} & m \\ \hline
    ${\hat \bfx}_0$ & $(125,125,125)^\top$ \rule{0cm}{0.01cm} & m \\ \hline
    $(n_1, n_2, n_3)$ & $(1,2,3)$ \rule{0cm}{0.01cm} & $---$ \\ \hline
    $\omega$ & $10^{-2}\pi$ \rule{0cm}{0.01cm} & rad/s \\ \hline
    $A_i n_i \omega$ (for $i=1,2,3$) & $0.5$ \rule{0cm}{0.01cm}  & m/s \\ \hline
    Sampling Time & $10^{-2}$ \rule{0cm}{0.01cm}  & s \\ \hline
    $Q$ & $10^{-4} I_{3 \times 3}$ \rule{0cm}{0.01cm}  & m$^2$ \\ \hline
    $R$  & $1$ \rule{0cm}{0.01cm}  & m$^2$. \\ \hline 
    \end{tabular}
\end{center} 
The resulting trajectories are depicted ad different zoom levels in figure \ref{figtrajnocurrent}: the initial position of the vehicle is marked with a green $*$ symbol (visible in the bottom plot). The vehicle trajectory is plotted in green while the Kalman estimate is plotted in red. The blue line is the one-step Kalman predicted position.   

\begin{figure}[thf]
\begin{center}
\includegraphics[width=8.0cm]{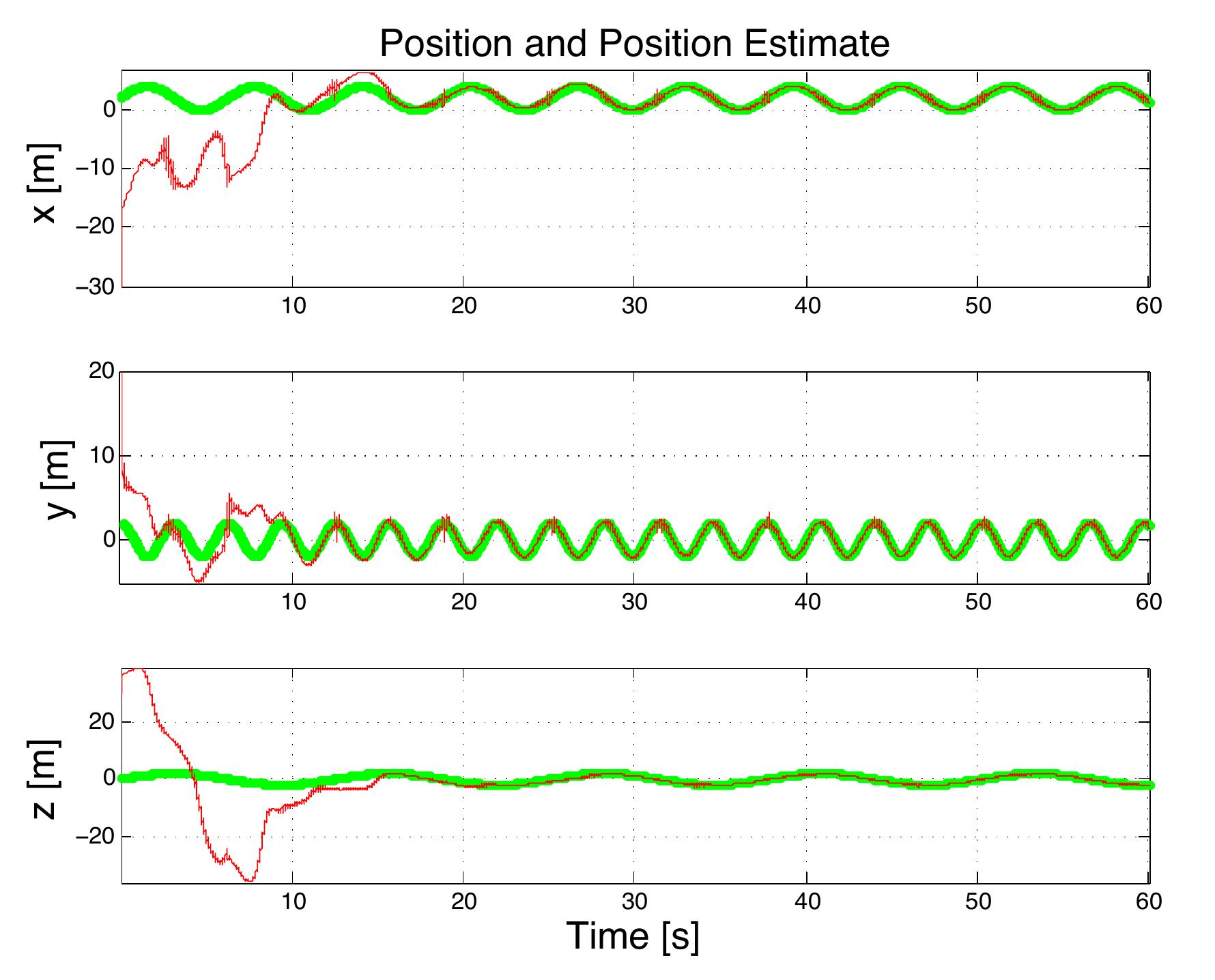}    
\caption{Kalman filter estimation including currents: real (in green) and estimated (in red) components of $\bfx$.} 
\label{figposwithcurrents}
\end{center}
\end{figure}

\begin{figure}[thf]
\begin{center}
\includegraphics[width=8.0cm]{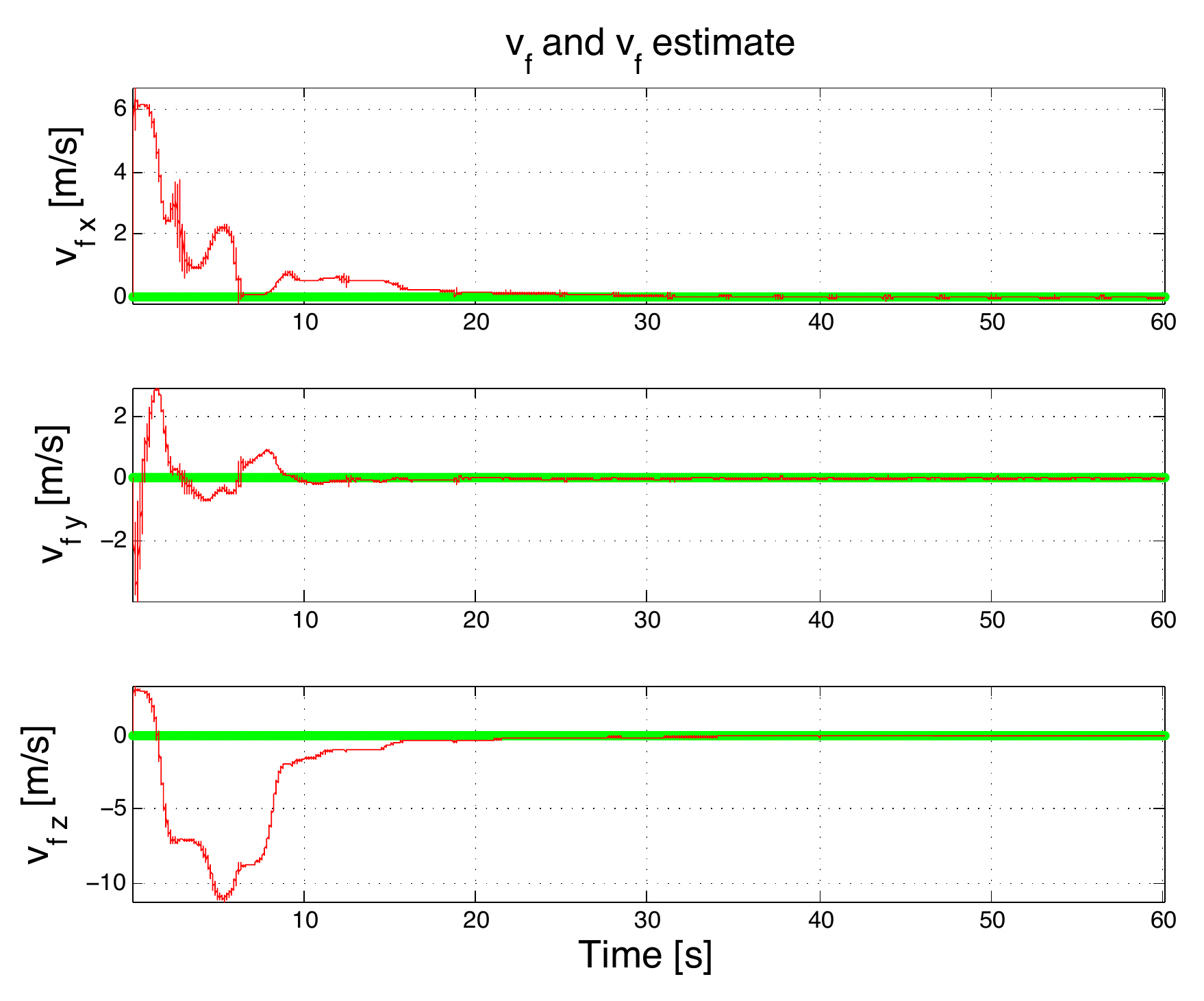}    
\caption{Kalman filter estimation including currents: real (in green) and estimated (in red) components of the current velocity $\bfv_f$.} 
\label{figCurrentestimate}
\end{center}
\end{figure}

As for the case presented in section \ref{sec:currents} in the presence of a constant and unknown current $\bfv_f$, the state $\bfz$ can be estimated with a Kalman filter. In particular, a numerical experiment is performed using the same agent velocity profile $\bfv_r$ used in the examples presented in \cite{DaFiDaAn:09} and \cite{Batista:11} namely 
$\bfv_r = (2\cos(t), -4\sin(2t), \cos(t/2))^\top$ (m/s). The target $\bfs$ is 
$\bfs = (2, 3, 1)^\top$ (m), the current is assumed null, $\bfv_f = {\bf 0}$ (m/s), and the initial position of the agent is $\bfx_0 = (2,2,0)^\top$ (m) such that the 
inertial position of the agent by $\bfx(t) = (2 + 2\sin(t), 2\cos(2t), 2\sin(0.5t))^\top$ (m). Notice that, by  direct calculation, the above $\bfv_r$ input satisfies the observability condition given in Statement 3.  The covariances on the state $\bfz$ and output ${\bar y}(t)$  employed in the Kalman filter are $Q=(1e-2)\mbox{diag}([1,1,1, 1e-4, 1e-6, (1e-2),(1e-2),(1e-2))$ and $R=1$ respectively with proper units (i.e. $[m^2]$ for position variables and $[(m/s)^2]$ for velocity variables). 
The filter is initialized with a position ${\hat \bfx}_0=(-30, 20, 30)^\top$[m] as opposed to the real initial position $\bfx_0 = (2,2,0)^\top$[m] and a current estimate ${\hat \bfv}_f=(0.1,-0.1,0.1)^\top$[m/s] as opposed to the real null current. Denoting with $T_s$ the sampling time (that was $(1/750)$[s] in the described example), the Kalman filter equations result in:
\bea
&& A_d = \left(I_{8 \times 8} + T_s\, A \right) \\
&& B_d = T_s\,B \\
&& {\hat \bfz}_{k+1|k} = A_d \, {\hat \bfz}_{k|k} + B_d\,{\bfv_r}_k \\
&& P_{k+1|k} =  A_d \, P_{k|k}\, A_d^\top + Q_k \label{predizionecov2}  \\
&& K = \left(P_{k+1|k}^{-1} + C_{k+1}^\top \, R_{k+1}^{-1} \, C_{k+1} \right)^{-1}C_{k+1}^\top \, R_{k+1}^{-1}\rule{0.85cm}{0.0cm} \\
&& {\hat \bfz}_{k+1|k+1} = {\hat \bfz}_{k+1|k} + K ({\bar y}_{k+1} - C_{k+1} \, {\hat \bfz}_{k+1|k} )
\eea
where $Q_k$ and $R_k$ were constant and equal to the values reported above.
 
The resulting  time evolution of the agent position $\bfx = \bfs -\bfr$ and its estimate ${\hat \bfx} = \bfs - {\hat \bfr}$ are plotted in figure \ref{figposwithcurrents} while the current estimate is plotted in figure \ref{figCurrentestimate}. 

\subsection{Discussion}
As already noticed, the proposed solution allows to design a Kalman filter for state estimation on a system where all the system matrices ($A$, $B$ and $C(t)$) are not affected by measurement  noise. This preserves the optimality of the Kalman filter as a state estimator in case of additive gaussian noise on the output and state equations. Yet the new output ${\bar y}(t)$ in equations (\ref{new-output-eq}) and 
(\ref{eq:barycurrentcase}) (and in their discrete time counterparts) always depends on the very first measurement $y(0)$. This dependency can impact on the robustness of the solution as a single bad measurement (as an outlier) at $t=0$ will affect the output for ever. 
A remedy to this issue can be found by periodically re-setting the initial measurement $y(0)$ with $y(t)$. In the discrete time case  this would correspond to periodically mapping 
$y_0 \; \longrightarrow \; y_{k^*}$ as if the measurement had started at step $k^*$ while the state estimate ${\hat \bfx_{k|k}}$ continues its update dynamics. A detailed analysis of this implementation detail goes beyond the scope of this paper and will not be addressed further, but it will be subject to future investigation. 

\section{Conclusions}
\label{sec:conclusions}
The problem of single range based localization for the kinematics model of a $3D$ vehicle was addressed in this paper. The problem is relevant in several filed robotics applications, particularly in underwater scenarios where ranges are measured acoustically and alternative radio frequency based localization devices as GPS are not available. Single range based localization techniques allow to avoid using trilateration based devices such as long base line (LBL) transponders that are very demanding in terms of cost and deployment effort.    
The vehicle is assumed to be equipped with standard on board navigation sensors as a doppler velocity logger DVL and an attitude heading reference system AHRS allowing to access the linear and angular vehicle velocities as well as the  system's attitude, i.e. the rotation matrix $^{I}R_B$ from body frame $\{B\}$ to the inertial frame 
$\{I\}$. 
The localization problem addressed is equivalent to the one presented in \cite{Batista:11} and it explicitly accounts for the effects of a constant, but unknown, ocean current that is estimated together with the vehicle position. 
The proposed solution allows to address the observability analysis and the state estimation filter design on a linear time invariant state equation defined on $\mathbb{R}^8$ with a time varying scalar output equation. In this respect the proposed solution resembles the one in \cite{Batista:11} where the original problem was transformed in a linear time varying state equation defined on $\mathbb{R}^9$ with a linear time invariant scalar output equation. Yet contrary to this previous solution, the state equation matrix $A$ does not depend on the inverse of the output $y(t)$ hence preserving the optimality of the Kalman filter in case of additive gaussian noise on the state and output equations. Moreover, the simple structure of the derived linear system for observability analysis allows to define straightforward necessary and sufficient observability conditions. The proposed solution can be applied in underwater cooperative navigation applications, sensor networks and source localization problems.  Examples of Kalman estimation filters for single range localization are provided in the absence and in the presence of constant ocean currents. 

\begin{ack}                               
This work was partially supported by MIUR (Italian Ministry of Education, Universities and Research) under the PRIN project MARIS: Marine Autonomous Robotics for InterventionS,  call of year 2010-2011, prot. 2010FBLHRJ.
\end{ack}

\bibliographystyle{plainnat}        
\bibliography{single-range-biblio}

\begin{thebibliography}{20}
\providecommand{\natexlab}[1]{#1}
\providecommand{\url}[1]{\texttt{#1}}
\expandafter\ifx\csname urlstyle\endcsname\relax
  \providecommand{\doi}[1]{doi: #1}\else
  \providecommand{\doi}{doi: \begingroup \urlstyle{rm}\Url}\fi

\bibitem[Arrichiello et~al.(2011)Arrichiello, Antonelli, Aguiar, and
  Pascoal]{Arrichiello:11}
Filippo Arrichiello, Gianluca Antonelli, Antonio~Pedro Aguiar, and Antonio
  Pascoal.
\newblock Observabiliy metrics for the relative localization of {AUV}s based on
  range and depth measurements: theory and experiments.
\newblock In \emph{Proceedings of the 2011 IEEE/RSJ International Conference on
  Intelligent Robots and Systems, IEEE-IROS 2011}, Hilton San Francisco Union
  Square, San Francisco, CA, USA, September 2011.
\newblock \doi{10.1109/IROS.2011.6094466}.
\newblock URL \url{http://dx.doi.org/10.1109/IROS.2011.6094466}.

\bibitem[Batista et~al.(2010)Batista, Silvestre, and Oliveira]{Batista:10}
Pedro Batista, Carlos Silvestre, and Paulo Oliveira.
\newblock Single beacon navigation: Observability analysis and filter design.
\newblock In \emph{Proceedings of the 2010 American Control Conference,
  ACC2010}, pages 6191--6196, Marriott Waterfront, Baltimore, MD, USA, June 30
  -- July 02 2010. IEEE.
\newblock ISBN 978-1-4244-7426-4.

\bibitem[Batista et~al.(2011)Batista, Silvestre, and Oliveira]{Batista:11}
Pedro Batista, Carlos Silvestre, and Paulo Oliveira.
\newblock Single range aided navigation and source localization: Observability
  and filter design.
\newblock \emph{Systems \& Control Letters}, 60:\penalty0 665--673, 2011.
\newblock \doi{10.1016/j.sysconle.2011.05.004}.
\newblock URL \url{http://dx.doi.org/10.1016/j.sysconle.2011.05.004}.

\bibitem[Cao et~al.(2011)Cao, Yu, and Anderson]{CaYuAn:11}
Ming Cao, Changbin Yu, and Brian~D.O. Anderson.
\newblock Formation control using range-only measurements.
\newblock \emph{Automatica}, 47:\penalty0 776--781, 2011.
\newblock \doi{10.1016/j.automatica.2011.01.067}.
\newblock URL \url{http://dx.doi.org/10.1016/j.automatica.2011.01.067}.

\bibitem[Dandach et~al.(2009)Dandach, Fidan, Dasgupta, and
  Anderson]{DaFiDaAn:09}
Sandra~H. Dandach, Bari\c{s} Fidan, Soura Dasgupta, and Brian~D.O. Anderson.
\newblock A continuous time linear adaptive source localization algorithm,
  robust to persistent drift.
\newblock \emph{Systems \& Control Letters}, 59:\penalty0 7--16, 2009.
\newblock \doi{10.1016/j.sysconle.2008.07.008}.
\newblock URL \url{http://dx.doi.org/10.1016/j.sysconle.2008.07.008}.

\bibitem[Fallon et~al.(2010)Fallon, Papadopoulos, Leonard, and
  Patrikalakis]{Fallon:10}
Maurice~F. Fallon, Georgios Papadopoulos, John~J. Leonard, and Nicholas~M.
  Patrikalakis.
\newblock Cooperative {AUV} navigation using a single maneuvering surface
  craft.
\newblock \emph{I. J. Robotic Res.}, 29\penalty0 (12):\penalty0 1461--1474,
  2010.
\newblock \doi{10.1177/0278364910380760}.
\newblock URL \url{http://dx.doi.org/10.1177/0278364910380760}.

\bibitem[Gadre and Stilwell(2004)]{Gadre:04}
A.~S. Gadre and D.~J. Stilwell.
\newblock Toward underwater navigation based on range measurements from a
  single location.
\newblock In \emph{Proceedings of IEEE International Conference on Robotics and
  Automation, 2004 (ICRA 2004)}, New Orleans, LA, USA, 26 April -- 1 May 2004
  2004.
\newblock \doi{10.1109/ROBOT.2004.1302422}.
\newblock URL \url{http://dx.doi.org/10.1109/ROBOT.2004.1302422}.

\bibitem[Hermann and Krener(1977)]{Hermann}
Robert Hermann and Arthur~J. Krener.
\newblock Nonlinear controllability and observability.
\newblock \emph{IEEE Transactions on Automatic Control}, 22\penalty0
  (5):\penalty0 728 -- 740, October 1977.
\newblock \doi{10.1109/TAC.1977.1101601}.
\newblock URL \url{http://dx.doi.org/10.1109/TAC.1977.1101601}.

\bibitem[Indiveri et~al.(2012)Indiveri, Pedone, and
  Cuccovillo]{InPeCuNGCUV2012}
Giovanni Indiveri, Paola Pedone, and Michele Cuccovillo.
\newblock Fixed target {3D} localization based on range data only: A recursive
  least squares approach.
\newblock In \emph{Proceedings of the 2012 IFAC Workshop on Navigation,
  Guidance and Control of Underwater Vehicles, IFAC-NGCUV2012}, pages 140 --
  145, Porto, Portugal, 10 - 12 April 2012.
\newblock \doi{10.3182/20120410-3-PT-4028.00024}.
\newblock URL \url{http://dx.doi.org/10.3182/20120410-3-PT-4028.00024}.

\bibitem[Jouffroy and Reger(2006)]{jouffroy2006algebraic}
J{\'e}r{\^o}me Jouffroy and Johann Reger.
\newblock An algebraic perspective to single-transponder underwater navigation.
\newblock In \emph{Computer Aided Control System Design, 2006 IEEE
  International Conference on Control Applications, 2006 IEEE International
  Symposium on Intelligent Control}, pages 1789--1794. IEEE, 2006.
\newblock \doi{10.1109/CACSD-CCA-ISIC.2006.4776912}.
\newblock URL \url{http://dx.doi.org/10.1109/CACSD-CCA-ISIC.2006.4776912}.

\bibitem[Kassas and Humphreys(2012)]{kassas2012observability}
Zaher~M Kassas and Todd~E Humphreys.
\newblock Observability analysis of opportunistic navigation with pseudorange
  measurements.
\newblock In \emph{AIAA Guidance, Navigation, and Control Conference, AIAA
  GNC}, 2012.
\newblock \doi{10.2514/6.2012-4760}.
\newblock URL \url{http://dx.doi.org/10.2514/6.2012-4760}.

\bibitem[Maki et~al.(2013)Maki, Matsuda, Sakamaki, Ura, and Kojima]{JOE-URA:13}
Toshihiro Maki, Takumi Matsuda, Takashi Sakamaki, Tamaki Ura, and Junichi
  Kojima.
\newblock Navigation method for underwater vehicles based on mutual acoustical
  positioning with a single seafloor station.
\newblock \emph{IEEE Journal Of Oceanic Engineering}, 38\penalty0 (1):\penalty0
  167 -- 177, 2013.
\newblock \doi{10.1109/JOE.2012.2210799}.
\newblock URL \url{http://dx.doi.org/10.1109/JOE.2012.2210799}.

\bibitem[Martinelli and Siegwart(2005)]{Martinelli:05}
A.~Martinelli and R.~Siegwart.
\newblock Observability analysis for mobile robot localization.
\newblock In \emph{Intelligent Robots and Systems, 2005. (IROS 2005). 2005
  IEEE/RSJ International Conference on}, pages 1471 -- 1476, aug. 2005.
\newblock \doi{10.1109/IROS.2005.1545153}.
\newblock URL \url{http://dx.doi.org/10.1109/IROS.2005.1545153}.

\bibitem[Parlangeli et~al.(2012)Parlangeli, Pedone, and
  Indiveri]{PaPeInMCMC2012}
Gianfranco Parlangeli, Paola Pedone, and Giovanni Indiveri.
\newblock Relative pose observability analysis for {3D} nonholonomic vehicles
  based on range measurements only.
\newblock In \emph{Proceedings of the 9th IFAC Conference on Manoeuvring and
  Control of Marine Craft, MCMC 2012}, Arenzano (GE), Italy, 19 - 21 September
  2012.

\bibitem[Ross and Jouffroy(2005)]{jouffroyremarks}
Andrew Ross and J{\'e}r{\^o}me Jouffroy.
\newblock Remarks on the observability of single beacon underwater navigation.
\newblock In \emph{Int. Symp. on Unmanned Untethered Submersible Technology
  (UUST 05)}, Durham, NH, August 2005.

\bibitem[Rugh(1996)]{Rugh:96}
Wilson~J. Rugh.
\newblock \emph{Linear System Theory}.
\newblock Prentice-Hall, 1996.
\newblock ISBN 0134412052.

\bibitem[Soares et~al.(2013)Soares, Aguiar, Pascoal, and
  Martinoli]{SoAgPaMa-ICRA2013}
Jorge~M. Soares, A.~Pedro Aguiar, Antonio~M. Pascoal, and Alcherio Martinoli.
\newblock Joint {ASV}/{AUV} range-based formation control: Theory and
  experimental results.
\newblock In \emph{2013 IEEE International Conference on Robotics and
  Automation ICRA 2013}, Karlsruhe, Germany, 6-10 May 2013.

\bibitem[Song(1999)]{Song:99}
Taek~Lyul Song.
\newblock Observability of target tracking with range-only measurements.
\newblock \emph{Oceanic Engineering, IEEE Journal of}, 24\penalty0
  (3):\penalty0 383 -- 387, jul 1999.
\newblock \doi{10.1109/48.775299}.
\newblock URL \url{http://dx.doi.org/10.1109/48.775299}.

\bibitem[Webster et~al.(2013)Webster, Walls, Whitcomb, and
  Eustice]{WeWaWhEu-TRO2013}
Sarah~E. Webster, Jeffrey~M. Walls, Louis~L. Whitcomb, and Ryan~M. Eustice.
\newblock Decentralized extended information filter for single-beacon
  cooperative acoustic navigation: Theory and experiments.
\newblock \emph{IEEE Transactions on Robotics}, \penalty0 (99):\penalty0 1--18,
  2013.
\newblock \doi{10.1109/TRO.2013.2252857}.
\newblock URL \url{http://dx.doi.org/10.1109/TRO.2013.2252857}.

\bibitem[Zhou and Roumeliotis(2008)]{zhou2008robot}
Xun~S Zhou and Stergios~I Roumeliotis.
\newblock Robot-to-robot relative pose estimation from range measurements.
\newblock \emph{IEEE Transactions on Robotics}, 24\penalty0 (6):\penalty0
  1379--1393, 2008.
\newblock \doi{10.1109/TRO.2008.2006251}.
\newblock URL \url{http://dx.doi.org/10.1109/TRO.2008.2006251}.

\end{thebibliography}




\end{document}